\documentclass[letterpaper, 10 pt, conference]{ieeeconf}  

\IEEEoverridecommandlockouts                              

\usepackage{epsfig} 
\usepackage[hidelinks]{hyperref}
\usepackage{xcolor}
\usepackage{tikz}
\usetikzlibrary{backgrounds}
\usepackage{pgfplots}

\usepackage[firstpage=true]{background}
\backgroundsetup{
    position=current page.south,
  angle=0,
  vshift=10mm,
  color=gray,
  scale=1,
  contents=IROS 2017 AGROB Workshop
}

\title{\LARGE \bf
From Plants to Landmarks: Time-invariant Plant Localization that uses Deep Pose Regression in Agricultural Fields
}
\author{ Florian Kraemer\hspace{15pt} Alexander Schaefer\hspace{15pt} Andreas Eitel\hspace{15pt} Johan Vertens\hspace{15pt} Wolfram Burgard
\thanks{All authors are with the Department of Computer Science at the University of Freiburg, Germany.
This work has partially been supported by the European Commission under the grant number H2020-ICT-644227-FLOURISH.
}
}

\usepackage{graphicx}
\usepackage[font=footnotesize]{caption}
\usepackage{svg}
\setsvg{inkscape=inkscape -z -D,svgpath=fig/svg/}
\usepackage{siunitx}
\usepackage{amsmath}
\usepackage{amssymb}
\usepackage[nodayofweek,level]{datetime}

\begin{document}
\newcommand{\anton}{A--W4} 			
\newcommand{\dora}{B--W5} 		 	
\newcommand{\gustav}{D--W4}			
\newcommand{\heinrich}{C--W8}			
\newcommand{\ludwig}{E--W5}			
\newcommand{\martha}{F--W6}			

\newcommand{\berta}{G--W5}			
\newcommand{\caesar}{H--W5}			
\newcommand{\emil}{I--W6} 			
\newcommand{\friedrich}{J--W8}			
\newcommand{\norbert}{K--W8}			

\newcommand{\svm}{SVM}
\newcommand{\segm}{Segm.}
\newcommand{\regr}{Regr.}

\maketitle
\thispagestyle{empty}
\pagestyle{empty}

\begin{abstract}
  Agricultural robots are expected to increase yields in a sustainable
  way and automate precision tasks, such as weeding and plant monitoring.
  At the same time, they move in a continuously changing,
  semi-structured field environment, in which features can hardly be
  found and reproduced at a later time.  Challenges for Lidar and
  visual detection systems stem from the fact that plants can be very small, 
  overlapping and have a steadily changing appearance.
  Therefore, a popular way to localize vehicles with high accuracy
  is based on expensive global navigation satellite systems and not on natural landmarks. The
  contribution of this work is a novel image-based plant localization
  technique that uses the time-invariant stem emerging
  point as a reference. Our approach is based on a fully convolutional
  neural network that learns landmark localization from RGB and NIR
  image input in an end-to-end manner. The network performs pose regression
  to generate a plant  location likelihood map. Our approach allows us
  to cope with visual variances of plants both for different species
  and different growth stages.  We achieve high localization
  accuracies as shown in detailed evaluations of a
  sugar beet cultivation phase.  In experiments with our BoniRob
  we demonstrate that detections can be robustly reproduced with
  centimeter accuracy. 
\end{abstract}

\section{INTRODUCTION}
The growth of the world population is predicted to not saturate until 2050 and its increasing demands must be met.
The agricultural output is expected to rise by \SI{60}{\%} while the available  farmland will experience competition from the growing interest in bio-fuels.
At the same time, modern agricultural systems are based on untargeted use of pesticides and inorganic nutrients.
Their use is already in question nowadays and some demand permacultures and less interference with biological systems that contribute naturally to the agricultural output \cite{bommarco2013ecological}.

The concept of Precision Farming might lower environmental impact while promising affordable, organic production with high yields.
Agricultural robots form the backbone of this system. They navigate the fields autonomously and intervene only in a targeted way.
For example, weeds are removed precisely by mechanical tools, and fertilizers can be placed selectively for individually monitored plants.

\begin{figure}[tbp]
\begin{minipage}[b]{.24\textwidth}
  \centering
  \includegraphics[width=\textwidth]{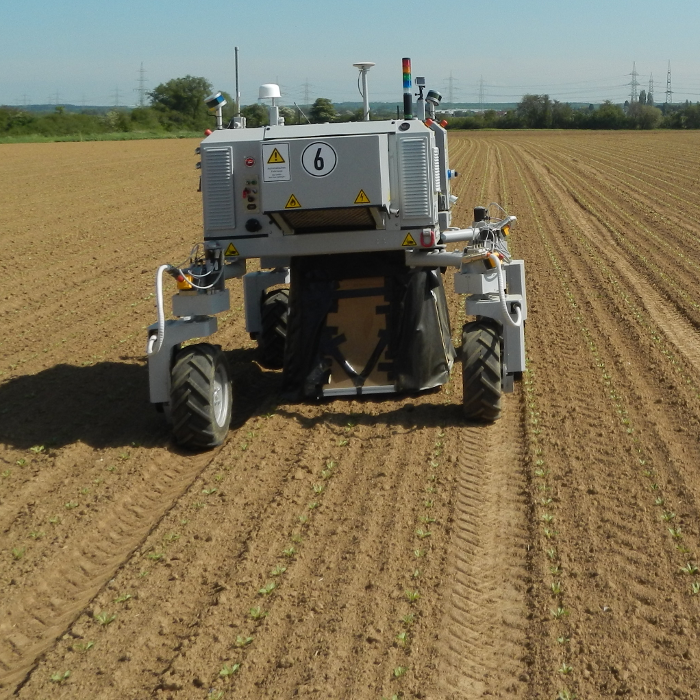}
  
\end{minipage}%
\hfill
\begin{minipage}[b]{.24\textwidth}
  \centering
  \includegraphics[width=\textwidth]{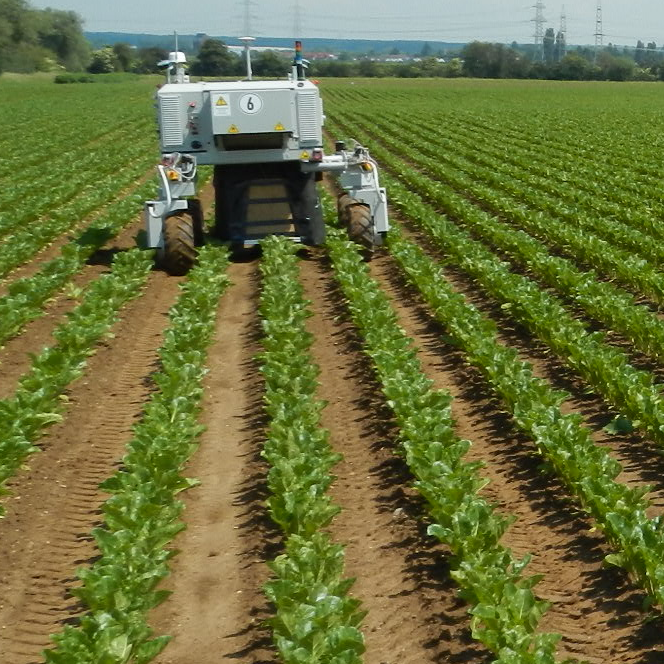}
\end{minipage}
  \caption{Flourish/Bonirob on a sugar beet field \SI{4}{weeks} and \SI{8}{weeks} after germination
  \cite{chebrolu2017agricultural}. The experimental setup is mounted under the robot, shields sunlight and records high-resolution RGB and NIR plant images.}\label{fig:flourish_early_to_late}
\end{figure}

However,  a mobile robot  must be able to localize itself in order to work autonomously.
This is particularly true for agricultural high-precision tasks.
Current agricultural robot implementations for fields rely on high-precision global navigation satellite systems (GNSS) to localize precisely. Examples are the system of Bakker et al. \cite{bakker2011autonomous}, the BoniRob setup without our extension \cite{ruckelshausen2009bonirob}, and AgBot/SwarmBot \cite{ball2017farm}, which can also localize temporarily by visual row tracking.
But this kind of GNSS requires the setup of a base station.
Additionally, poor signals due to forest borders and at longer distances from the base station can lower the accuracy.
An alternative to GNSS is that the robot maps its surrounding and localizes itself based on reproducible features.
The semi-structured and changing field environment, depicted in Fig.~\ref{fig:flourish_early_to_late}, poses a  difficult situation when not relying on artificial landmarks.
Here, it is highly challenging to reproduce standard image and point cloud features after some days.
Therefore, this work proposes time-invariant plant localization from multi-spectral images to lower the dependency on GNSS systems.
The estimated locations can be used in robotic mapping and localization approaches as landmarks and allow the robot to pursue its precision tasks.

\begin{figure}[tbp]
  \centering 
  \includegraphics[width=.48\textwidth]{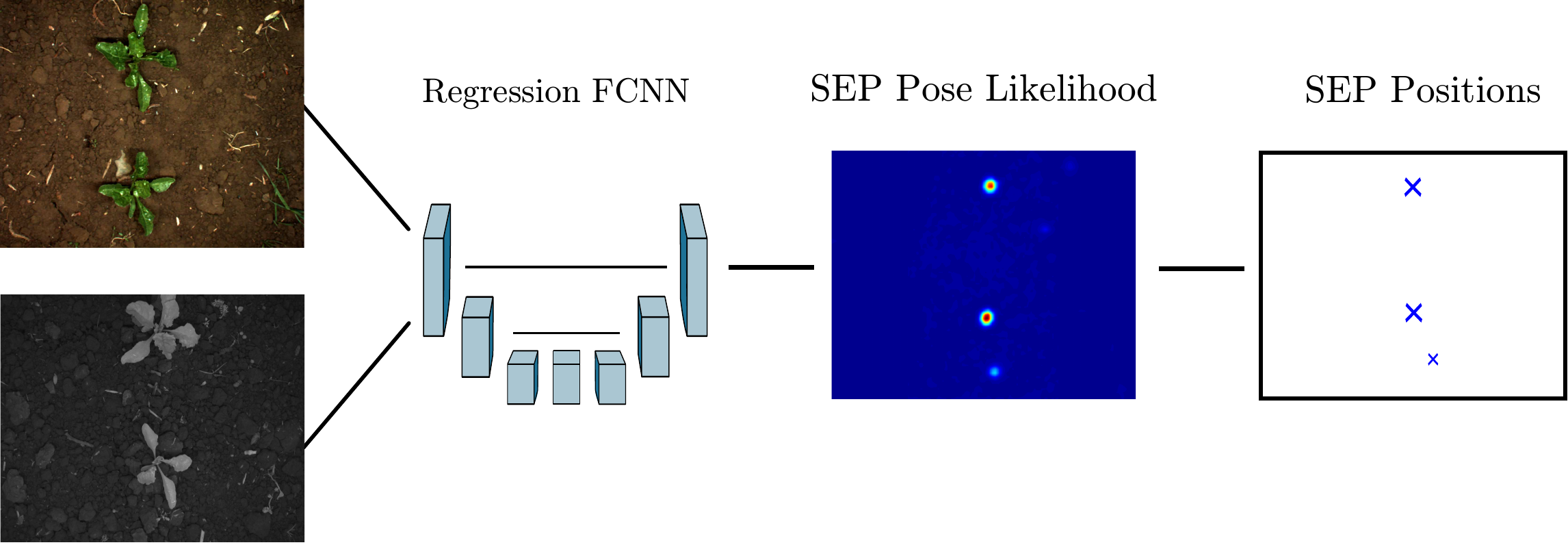}
  \caption{Deducing the Stem Emerging Points (SEPs) from RGB and NIR input images: We train a fully convolutional neural network (FCNN) to predict a pose likelihood map, from which we extract the SEP positions.}
  \label{fig:intro:covergirl}
\end{figure}

The key idea for the time-invariant plant localization is that the point of plant emergence never changes.
To detect the Stem Emerging Points (SEPs) in images we use a fully convolutional neural network to generate a pose likelihood map, as depicted in Fig.~\ref{fig:intro:covergirl}.
From the pose likelihood map, we extract SEPs for all plant species occurring in sugar beet fields and for different plant growth stages.

\begin{figure}
\begin{minipage}[b]{0.24\textwidth}
\centering
  \includegraphics[width=\textwidth]{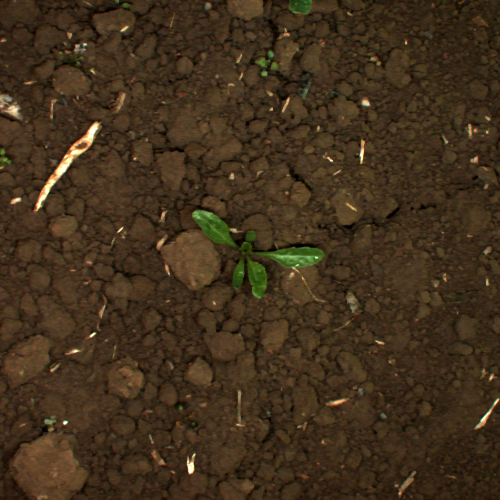}
\footnotesize{  \textbf{a)} Young, well separated sugar beet}
  \end{minipage}%
\hfill
\begin{minipage}[b]{0.24\textwidth}
  \centering
    \includegraphics[width=\textwidth]{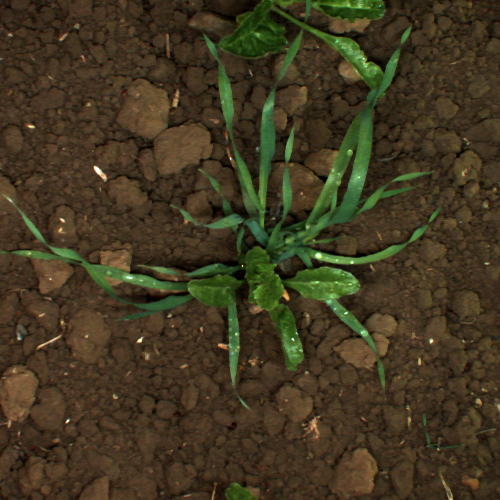}
    \footnotesize{  \textbf{b)} Intruding grass weed}
\end{minipage}%

\vspace{2pt}

\begin{minipage}[b]{0.24\textwidth}
  \centering
  \includegraphics[width=\textwidth]{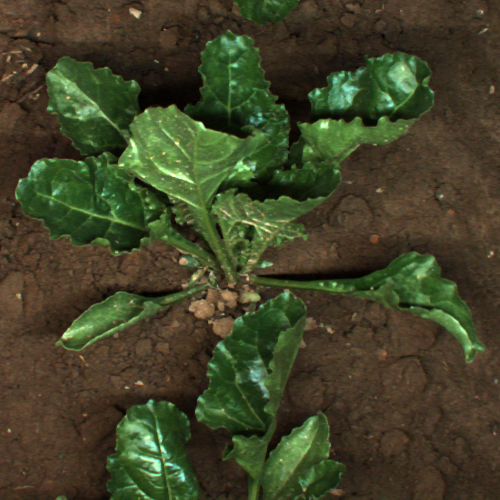}
  \footnotesize{  \textbf{c)} Deformation producing side view}
\end{minipage}%
\hfill
\begin{minipage}[b]{0.24\textwidth}
  \centering
    \includegraphics[width=\textwidth]{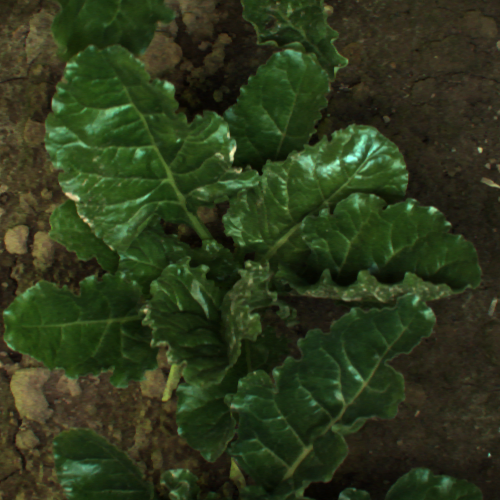}
    \footnotesize{  \textbf{d)} Overlaps at later growth stages}
\end{minipage}%
 \caption{Four examples of typically observed plants: a) and b) are taken \SI{6}{weeks}, c) \SI{7}{weeks} and d) \SI{8}{weeks} after germination.}
  \label{fig:intro:samples}
\end{figure}

Sugar beets, among other root vegetables like carrots, onions, or radish,  offer a good experimental environment for agricultural robots, since they are planted in well-traversable rows. 
Sugar beets grow near the ground and must be monitored for a period of six to eight weeks until the field enters the pre-harvesting phase \cite{chebrolu2017agricultural}. 
The field may contain severe plant overlaps and different weed species, as shown in example field images in Fig. \ref{fig:intro:samples}.
To account for difficult real world scenarios, we extract plant locations not only for sugar beet plants, but also for weeds present in these fields.
This allows including all plants in mapping and localization tasks, as well as targeting weed removal actions more precisely.
Our evaluation covers the full evolution of a sugar beet field observed from a ground robot. It shows the SEP localization accuracy for different growth stages and weed densities.

We make the following contributions: 1) A  SEP localization approach based on pose regression and fully convolutional neural networks, 2) a comprehensive evaluation on hand-annotated image data with several detection measures focusing on long-term robustness and comparing the novel approach with two baseline approaches, 3) a comparison of landmark maps generated by our BoniRob, showing the reliability of SEP landmarks over time, and 4) a dataset for plant localization and evaluation tools, available at:\\
\url{http://plantcentroids.cs.uni-freiburg.de}.

\section{RELATED WORK}\label{rel_work}
Plant localization is useful for agricultural robots in a number of ways.
In orchards, the semi-structured environment has been used for mapping and localization purposes: Cheein et al. \cite{cheein2011eifslam} find olive tree stems by camera and laser scanner detection, which then work as landmarks in a SLAM system.

Several approaches exist for individual field plant detection and SEP localization, many of which are highly use-case adapted and based on heuristics.
Midtiby et al. \cite{midtiby2012estimating} follow sugar beet leaf contours to find the SEPs.
This works only for very early growth stages without overlapping leaves.
Jin et al. \cite{jin2009corn} focus on corn plants, but use RGB-D data.
Again, they make strong model assumptions and follow the plants' image skeletons towards minima in the height domain.
Weiss and Biber \cite{weiss2011plant} segment plants based on point cloud clusters and estimate the plant positions at the median points of the clusters.
Since this might fail for overlapping plants, Gai et al. \cite{gai2016plant} extend this approach and follow leaf ridges detected in RGB-images to the center.
Their approach only works for specific crop types and their ground-plant segmentation fails for early growth stages.
Most similar to our work, Haug et al. \cite{haug2014stem} perform machine learning-based SEP localization in an organic carrot field.
They classify key points sampled equally-spaced from plant areas in NDVI images.
As a classifier, they employ a Random Forest with hand-crafted statistical and geometrical features for different patch sizes.
Their evaluation over \SI{25}{images} takes one point in time into account and they find weaknesses of the classifier for plant overlaps and locally misleading plant shapes.
Our work overcomes these by supplying a broader field of view to the classifier and by avoiding hand-crafted features that might only work for limited scenarios and plant species.

The field of image landmark localization is important to our pose regression approach.
Tompson et al. \cite{tompson2014real} estimate likelihood maps of hand joint locations based on depth images.
However, without any upsampling operation in their network architecture, they generate a coarse output of less than a fifth of the input size.
Zhang et al. \cite{zhang2015fine} overcome this drawback by using a fully convolutional neural network (FCNN). 
This allows them to generate likelihood maps with input image resolution.

\section{APPROACH}\label{approach}
We localize the SEPs in multi-spectral images by solving a pose regression problem with a neural network. 
FCNNs can learn in an end-to-end manner to output a pose likelihood map with high-resolution output.
FCNNs generally feature a contracting path similar to commonly used image classification networks.
In addition, FCNNs upsample the coarsely-resoluted bottleneck feature maps and localize features in an expanding path \cite{long2015fully}.
Formally, the problem is to assign every pixel a likelihood $p:\textbf{x}\mapsto{\rm I\!R}$ depending on the distance to the next SEP location.
We generate the ground truth values as $r:\textbf{x}\mapsto[0,1]$ to obtain a normalized network output that allows postprocessing it as likelihood maps.

In order to compute the pixel-wise ground truth regression values $r$, we annotate the plant locations for every input image by hand, following a set of rules:
For rosette-type plants, like sugar beet, the SEPs can be determined efficiently by following the leaf stems.
For herbaceous plants, like grass weeds, the region of emergence can be arbitrarily shaped.
In these cases, the emergence region is marked with a polygon contour. 
A post-processing step then estimates the SEP coordinates as the region's center of mass. 
Cases in which the human expert is uncertain are filtered out.

\begin{figure}
  \centering
    \includegraphics[width=.48\textwidth]{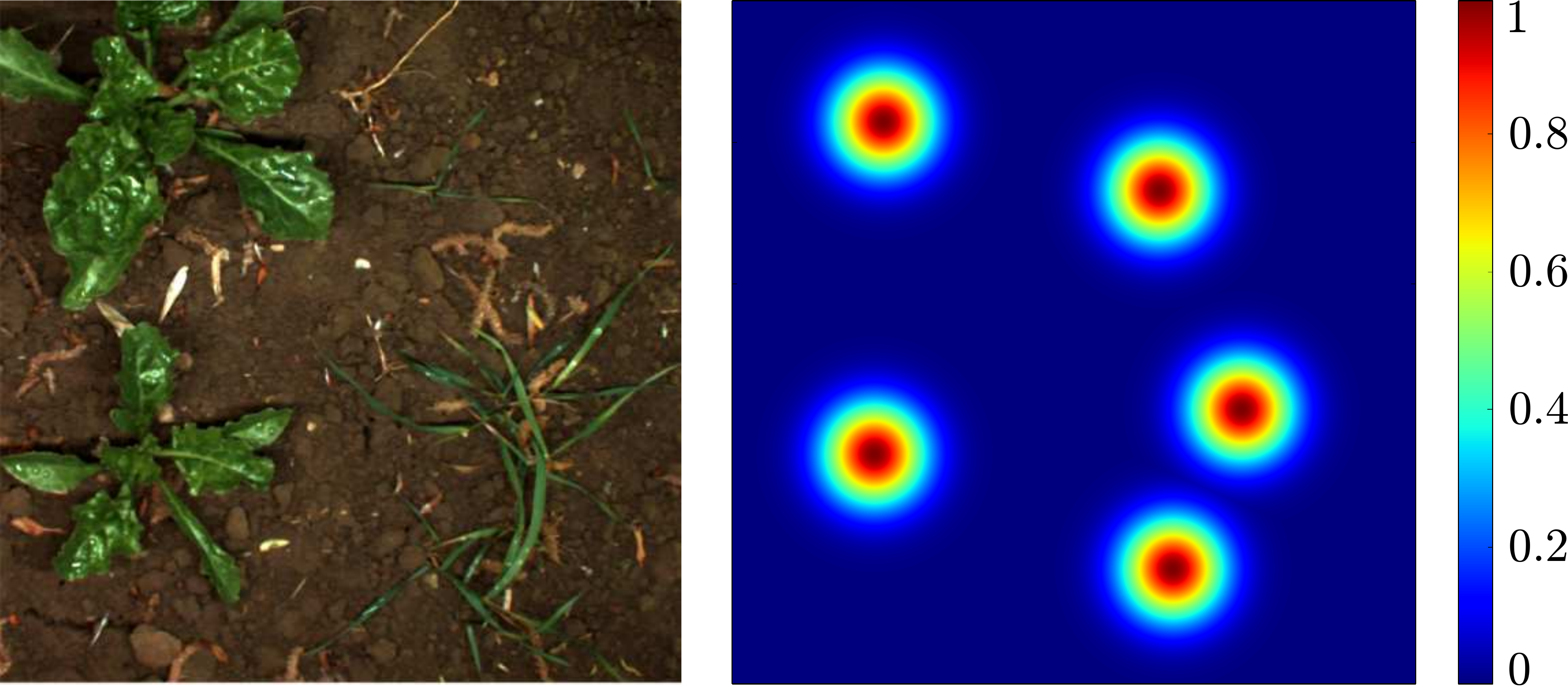}
 \caption{An example of the likelihood map ground truth generation taken from the sixth week after sugar beet germination. The RGB image (left) shows two sugar beet and several grass weed plants. The likelihood values (right) have been computed with $\sigma = \SI{50}{px}$ for visualization purposes.}
  \label{fig:approach:regr_example}
\end{figure}

From the ground truth SEP coordinates in every image set, we generate a likelihood map that is used for training the network.
First, a distance transform produces a map of distances to the closest ground truth SEP position $\textbf{x}_\textrm{SEP}$:
\begin{eqnarray}
d(\textbf{x})=\min\limits_{\textbf{x}_\textrm{SEP} \in \textrm{GT}} \lVert\textbf{x} -\textbf{x}_\textrm{SEP}\rVert_2
\end{eqnarray}
Next, we transform the resulting distance map to a Gaussian ground truth likelihood map using

\begin{eqnarray}
r(\textbf{x}) = \exp\frac{-d(\textbf{x})^2 }{2\sigma^2},
\end{eqnarray}
\noindent where $\sigma$ is a parameter that determines the decay of likelihood with increasing distance.
Fig. \ref{fig:approach:regr_example} shows a corresponding likelihood map.
We use the validation image set to tune $\sigma$ for the best precision-recall relation.
In our use-case, we achieve good results for $\sigma$ values from \SIrange{15}{19}{px}.

We use the L2-Norm, or Euclidean distance, of the network's regression score $p$ and the ground truth $r$ to compute the loss during network training.
The weight optimization by stochastic gradient descent can accordingly be formulated as minimizing the energy function for all pixel locations $\Omega$:

\begin{eqnarray}
\textrm{E} = \sum\limits_{\textbf{x} \in \Omega} \lVert r(\textbf{x}) - p(\textbf{x}) \rVert_2^2
\end{eqnarray}

Solving the pose regression problem does not depend on a specific fully convolutional neural network architecture.
Instead, architectures commonly used for semantic segmentation problems can be adapted to our problem.
We employ the architecture of MultiNet (the expert network from \cite{valada2017adapnet}), which has been successfully deployed with the restrictions of mobile robotics.
The low number of parameters, when compared to e.g. FCN8 \cite{long2015fully}, fits the memory capabilities of current embedded devices and allows processing our image stream in real-time at about \SI{4}{Hz}.
Additionally, the number of parameters hinders overfitting, when working with small data sets.
MultiNet extends the ResNet50 architecture in several domains:
The first feature maps are kept at a higher resolution since they need to be reused in the expanding path later on.
Additionally, it extends the residual blocks to two parallel convolutions, one of which has a wider kernel, to learn multi-scale features.
We use a Euclidean loss layer and implement the neural network in the Caffe framework \cite{jia2014caffe}.
The image input in the experiments is downsampled by a factor of \num{2} to capture plants entirely.
We stack the RGB and NIR images channel-wise and crop them to fit the network input size of $4 \times 512 \times 512$ (channels$\times$width$\times$height).

\begin{figure}
\begin{minipage}[b]{0.24\textwidth}
  \centering
  \includegraphics[width=\textwidth]{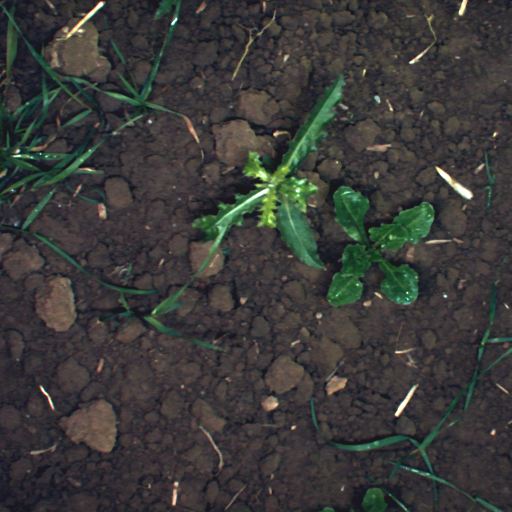}
  \end{minipage}%
\hfill
\begin{minipage}[b]{0.24\textwidth}
  \centering
    \includegraphics[width=\textwidth]{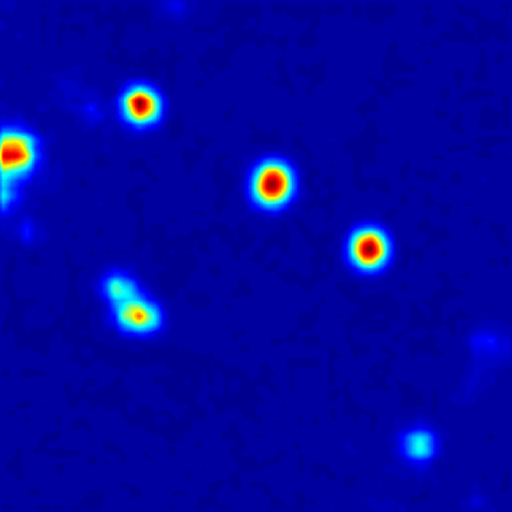}
\end{minipage}%

\vspace{3pt}

\begin{minipage}[b]{0.24\textwidth}
  \centering
    \includegraphics[width=\textwidth]{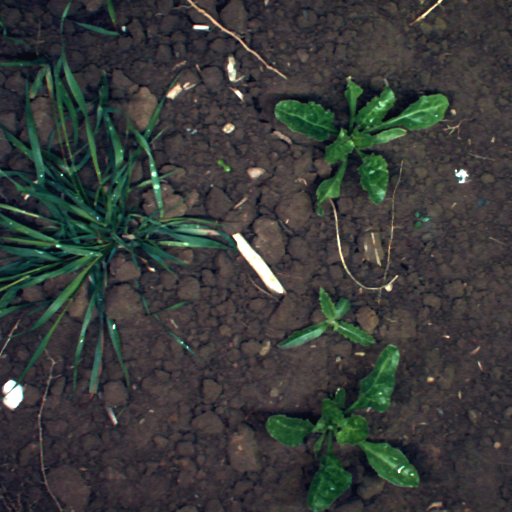}
    \end{minipage}%
  \hfill
  \begin{minipage}[b]{0.24\textwidth}
    \centering
      \includegraphics[width=\textwidth]{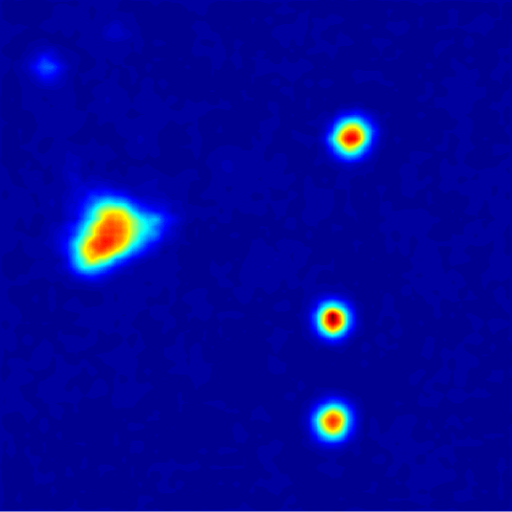}
\end{minipage}%
 \caption{Trained network likelihood map output (right), RGB image depicted for input visualization (left): Big grass weeds produce spread-out likelihood peaks, which are more spiked for rosette-type plants. Even hidden, small grass weeds are detected (see bottom image, upper left corner).}
  \label{fig:approach:examples}
\end{figure}

Examples for a trained network's output can be seen in Fig.~\ref{fig:approach:examples} where the likelihood map contains flat basins for regions distant to SEPs.
Furthermore, high likelihood regions correspond to the Gaussian likelihoods given in the ground truth, but they are usually more elongated.
For herbaceous plants, the peaks can be arbitrarily shaped. 
Generally, the peaks reach different likelihood values.
Their height can be seen as an indicator for the network's confidence or pseudo-probability for a detection.

The final step of our SEP localization approach is to extract the SEP coordinates from the likelihood map.
First, we segment the likelihood map into basins and peak regions using Otsu's method.
SEP locations are then determined as the center of mass for every likelihood peak region.
The confidence score for a detection is deduced as a region's average likelihood.
Scoring by the average value of a peak region reflects that detections are less confident when the likelihood peak is spread out. 
This way of confidence scoring performs better in our tests than relying on a peak region's maximum likelihood value.

\section{EXPERIMENTS}\label{results}
First, we compare our pose regression approach with two baselines using a hand-annotated field image test data set.
Then, we evaluate the reliability of the our approach, when plant locations are used as landmarks in a field mapping task.

As a first baseline, we re-implement the approach of Haug et al. \cite{haug2014stem}.
We found that it copes better with the big variances in our dataset when the feature set and classifier are adapted.
Instead of hand-crafted features, we use HOG-features of NDVI images.
We sample the key points with a stride of \SI{10}{px} and derive features by using patch sizes of \SI{40}{px}, \SI{80}{px}, \SI{160}{px} and \SI{320}{px}.
The classification of every key point's features is performed by an SVM with a radial basis function kernel.

As a second baseline, we extend the first baseline to also show the performance of a state-of-the-art classifier in this scenario.
Here, instead of coarsely sampled key points, we use the pixel-wise classifications of a segmentation FCNN.
The FCNN learns to classify for every pixel whether it belongs to the {Background}, {Vegetation} or {Stem Emerging Region} class.
The Vegetation class is introduced to enable the learning of richer features through finer class partitioning.
Also, the intermediate step of estimating regions of stem emergence rather than one stem emerging \textit{point} per plant has been introduced to keep class imbalances low.
The SEPs are then determined in a post-processing step as the center of mass for a given stem emerging connected component in the segmentation mask.
We estimate a confidence score for every region/SEP as the mean of pixel-wise network score differences between the stem emerging and the other classes.
In contrast to our regression approach, here, we annotate pixel-wise to generate ground truth segmentation masks.

\subsection{Test Image Set Comparison}
In this experiment, we evaluate the two baselines and our approach, using hand-annotated ground truth images.
All data is sampled from the Flourish Data Acquisition Campaign 2016 \cite{chebrolu2017agricultural} (see Fig. \ref{fig:flourish_early_to_late}).
Among other sensors, it features high-precision GNSS, odometry and a multi-spectral camera.
The camera records high-resolution image streams of the traversed sugar beet rows with an RGB spectrum of \SIrange{400}{680}{nm} and an NIR spectrum of \SIrange{730}{950}{nm}.
We use the aligned image channels to create a dataset.
Accordingly, we label several sequences of about \SI{300}{images} each, taken from different rows and on different days.
We denote them X--Wy where X is a unique identifier and y the number of weeks after sugar beet germination.
We use entire sequences to split the dataset into training and test partitions, since the images sequences are locally correlated.

To evaluate the output of an approach, we compute the distance $\textrm{d} = \min_{i\in\mathrm{GT}} \lVert \textbf{x}_i - \textbf{x} \rVert_2$ between the approaches' SEP detections and the ground truth SEP locations. 
Following standards in detection evaluation, we sort them by their confidence score in descending order.
Subsequently, we check whether the distance error lies below a threshold and only accept the highest scored detection per ground truth label.
The threshold is varied from \SIrange{6}{18}{px}, which corresponds to a ground resolution of about \SIrange{3,6}{8,4}{mm} at the image nadir (neglecting ground unevenness).
The binary result of being accepted or not allows us to use the quantitative measures of precision and recall.
In order to depict the distance errors better, we compute their mean for all accepted detections, the mean accepted distance (MAD), and associate it with the Average Precision (AP).
In this case, we accept a detection if it is closer than the minimal plant distance in the dataset, \SI{20}{px}, which we see as a natural boundary for acceptance.

All approaches of this comparison need to be trained.
The SVM baseline is trained on the \caesar{} sequence only, since more samples make the fitting process problematic.
The segmentation (second baseline) and regression networks are trained on the \berta{}, \caesar{}, \emil{}, \friedrich{} and \norbert{} sequences, which add up to \SI{1398}{images}.
They cover four weeks of sugar beet field evolution and have a varying weed content.
We augment the image set by systematically sampling 8 {rotated} and 2 {mirrored} versions as well as 4 randomly {cropped} versions of the input images.
Then, we train both networks for \SI{50000}{iterations} and a mini-batch size of 4 from scratch.

\begin{figure}[tp]
\begin{minipage}[b]{.24\textwidth}
  \centering  
    \anton{} \SI{6}{px}
   \includegraphics[width=\textwidth]{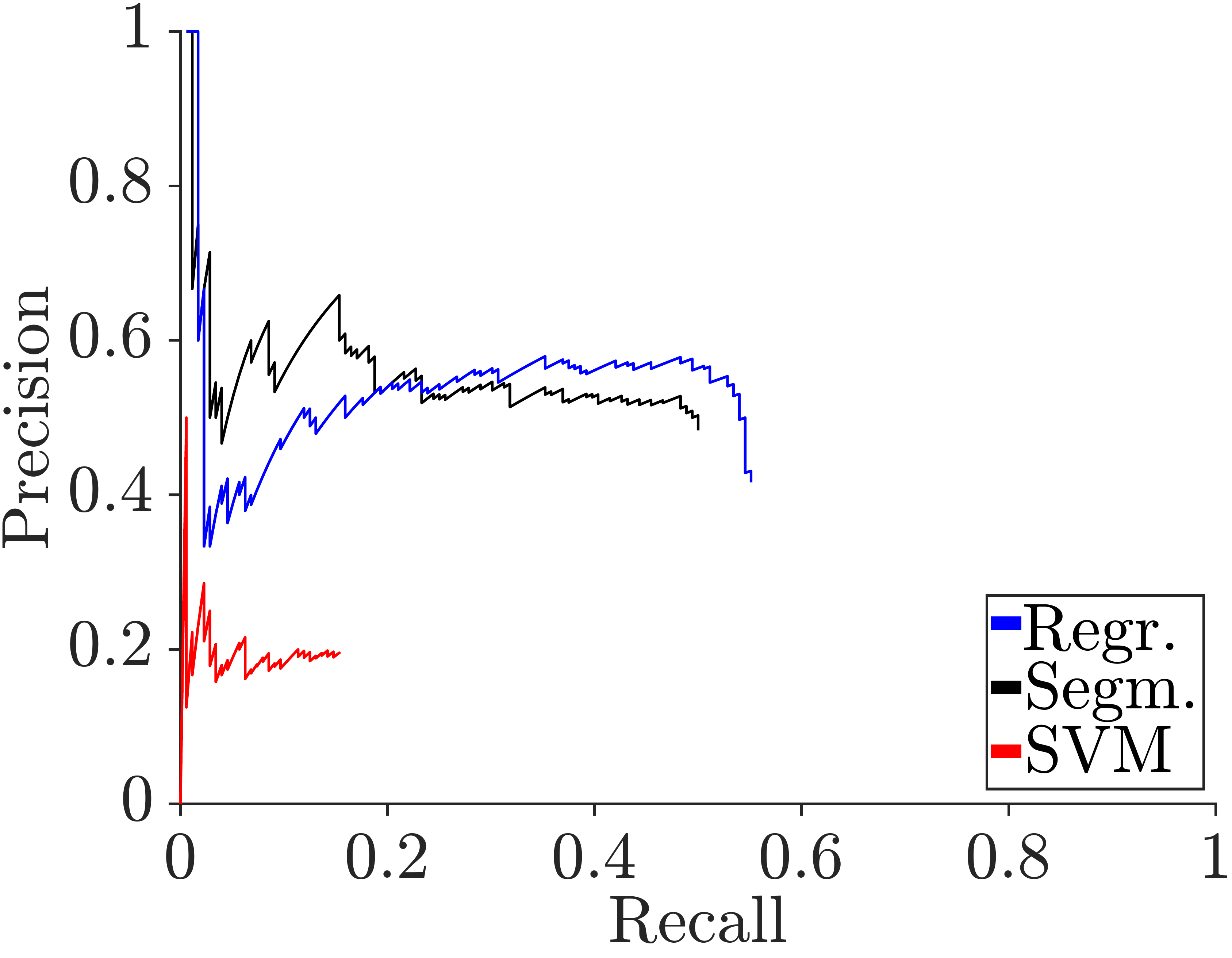}
\end{minipage}%
\hfill
\begin{minipage}[b]{.24\textwidth}
 \centering
   \anton{} \SI{18}{px}
  \includegraphics[width=\textwidth]{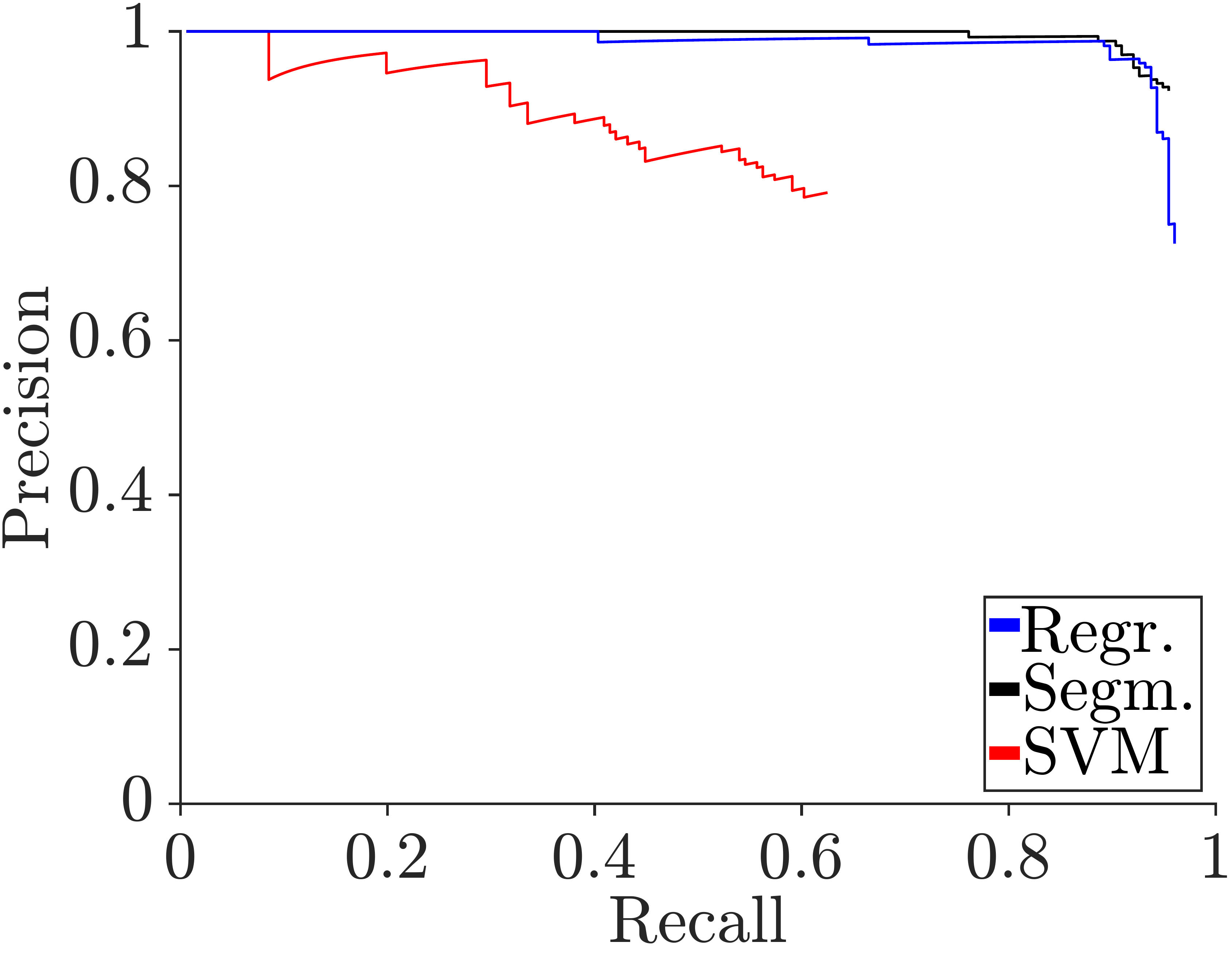}
\end{minipage}

\vspace{10pt}

\begin{minipage}[b]{.24\textwidth}
  \centering  
    \dora{} \SI{6}{px}
   \includegraphics[width=\textwidth]{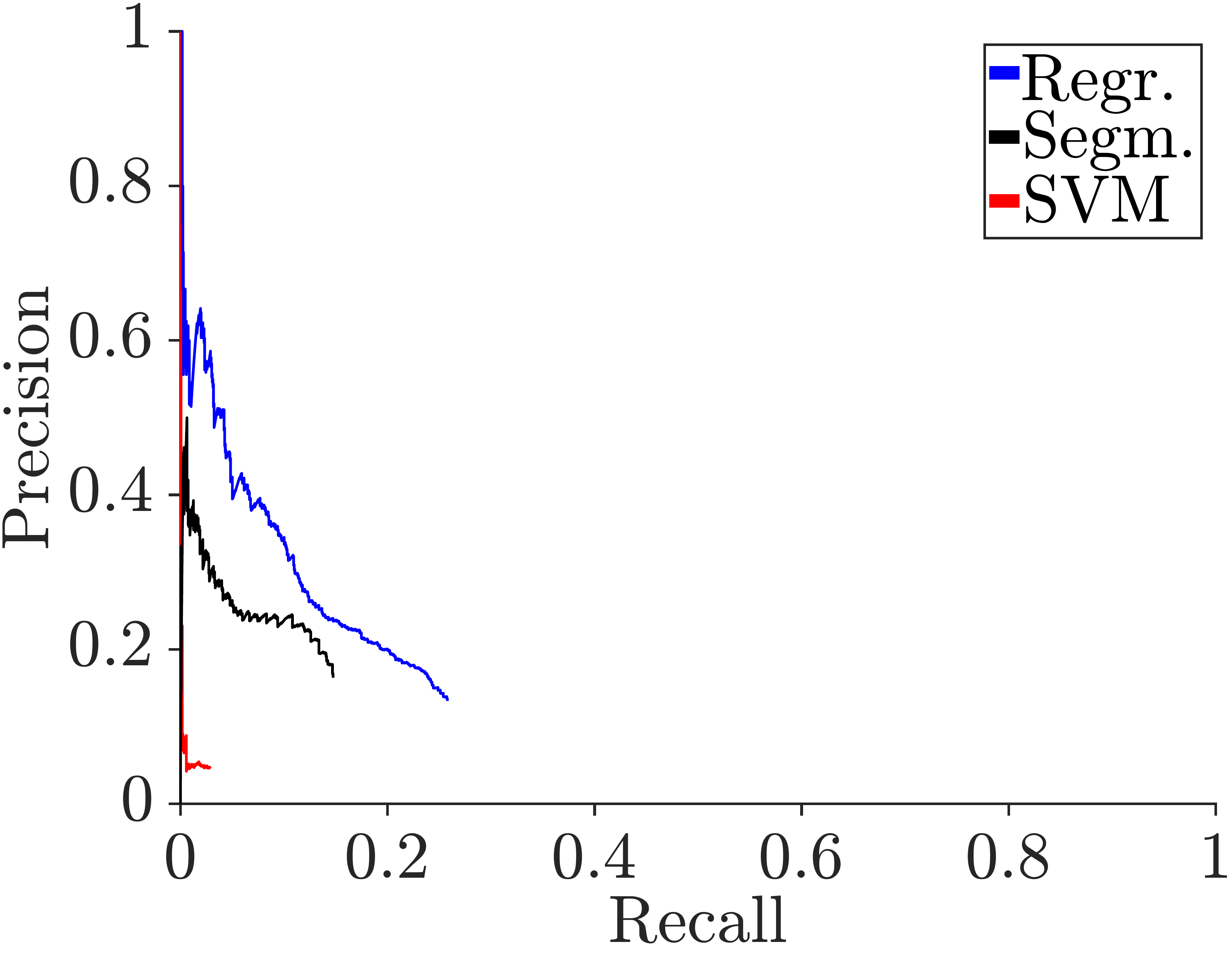}
\end{minipage}%
\hfill
\begin{minipage}[b]{.24\textwidth}
 \centering
     \dora{} \SI{18}{px}
  \includegraphics[width=\textwidth]{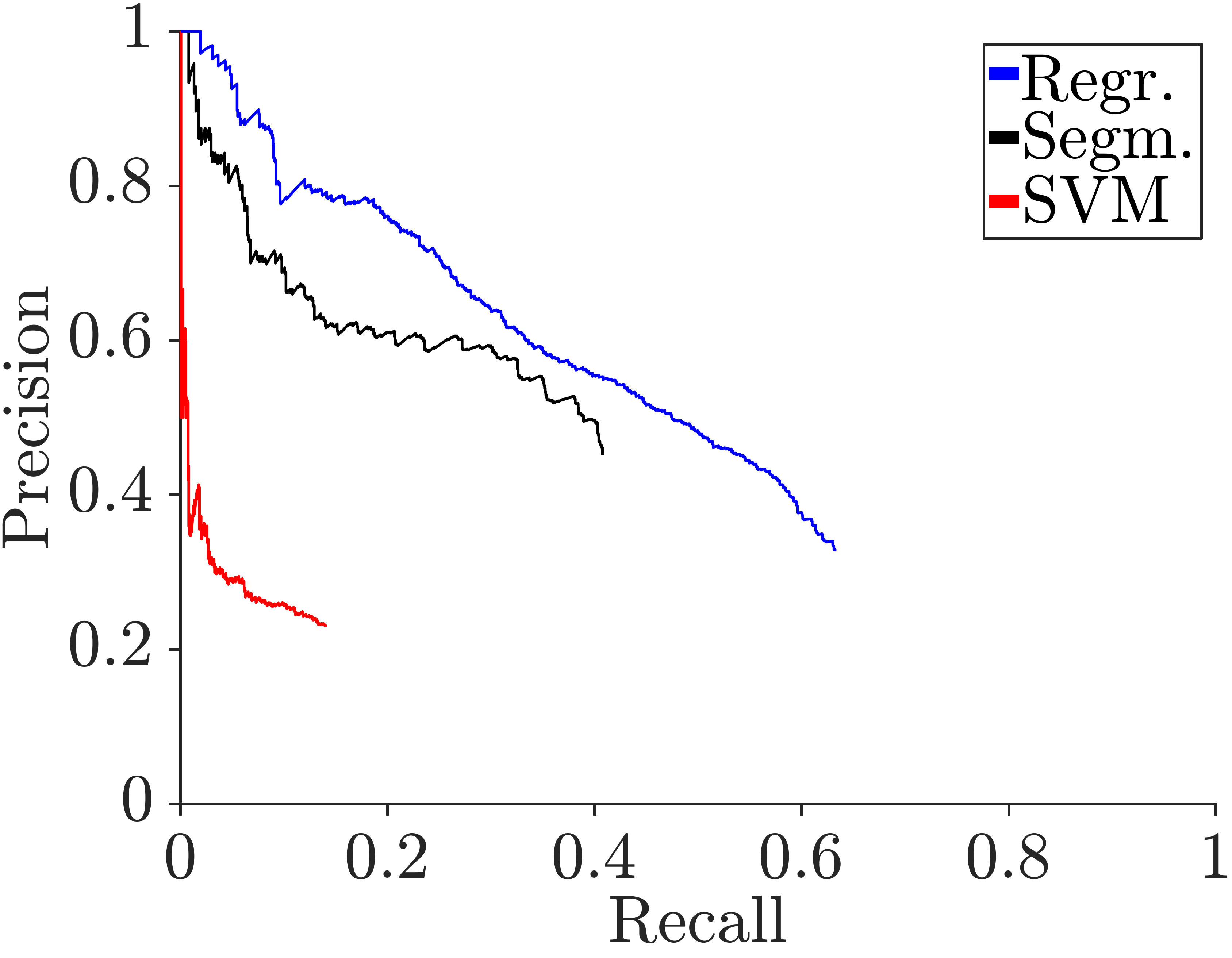}
\end{minipage}

\vspace{10pt}

\begin{minipage}[b]{.24\textwidth}
  \centering
      \heinrich{} \SI{6}{px}
     \includegraphics[width=\textwidth]{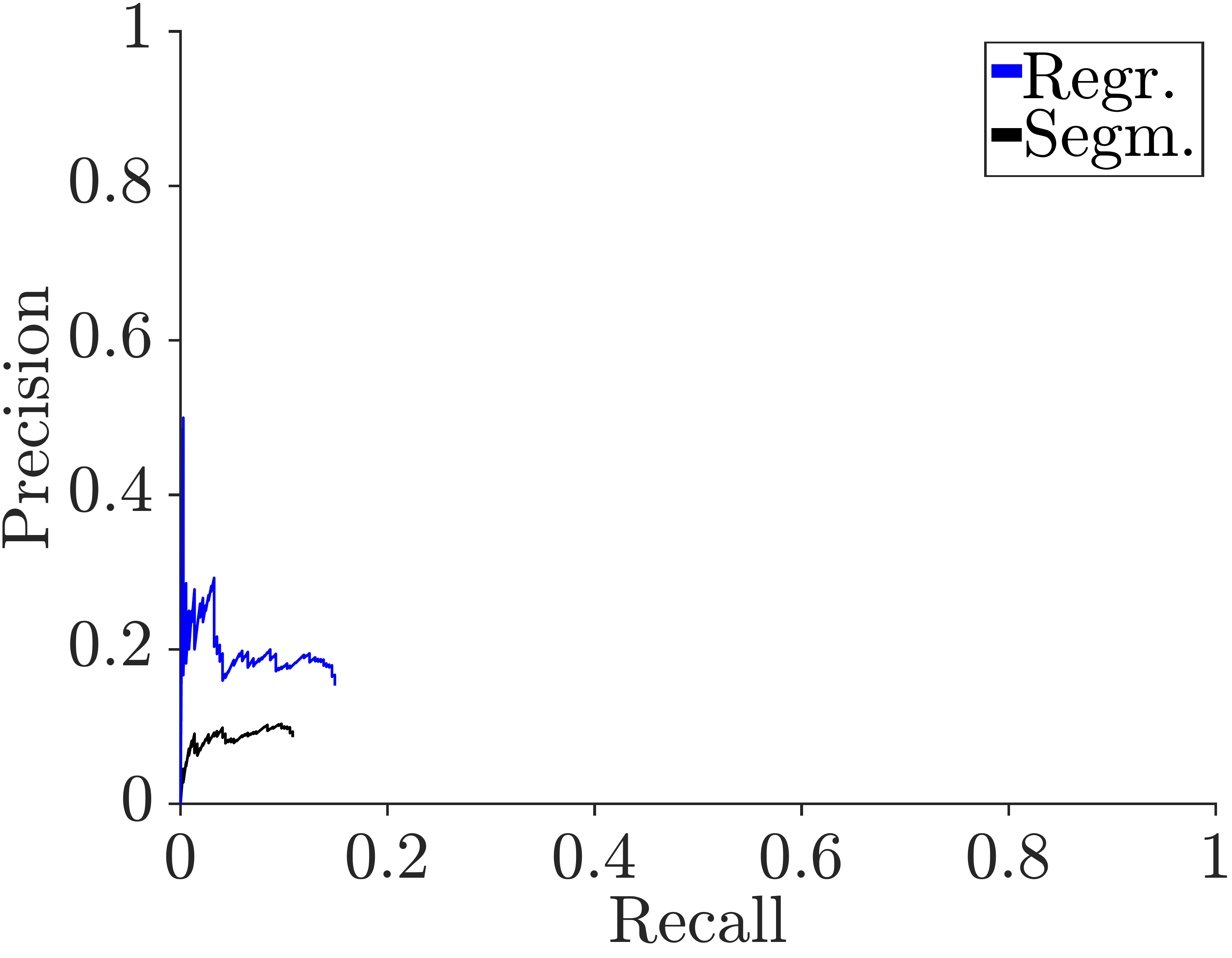}
\end{minipage}%
\hfill
\begin{minipage}[b]{.24\textwidth}
 \centering
       \heinrich{} \SI{18}{px}
    \includegraphics[width=\textwidth]{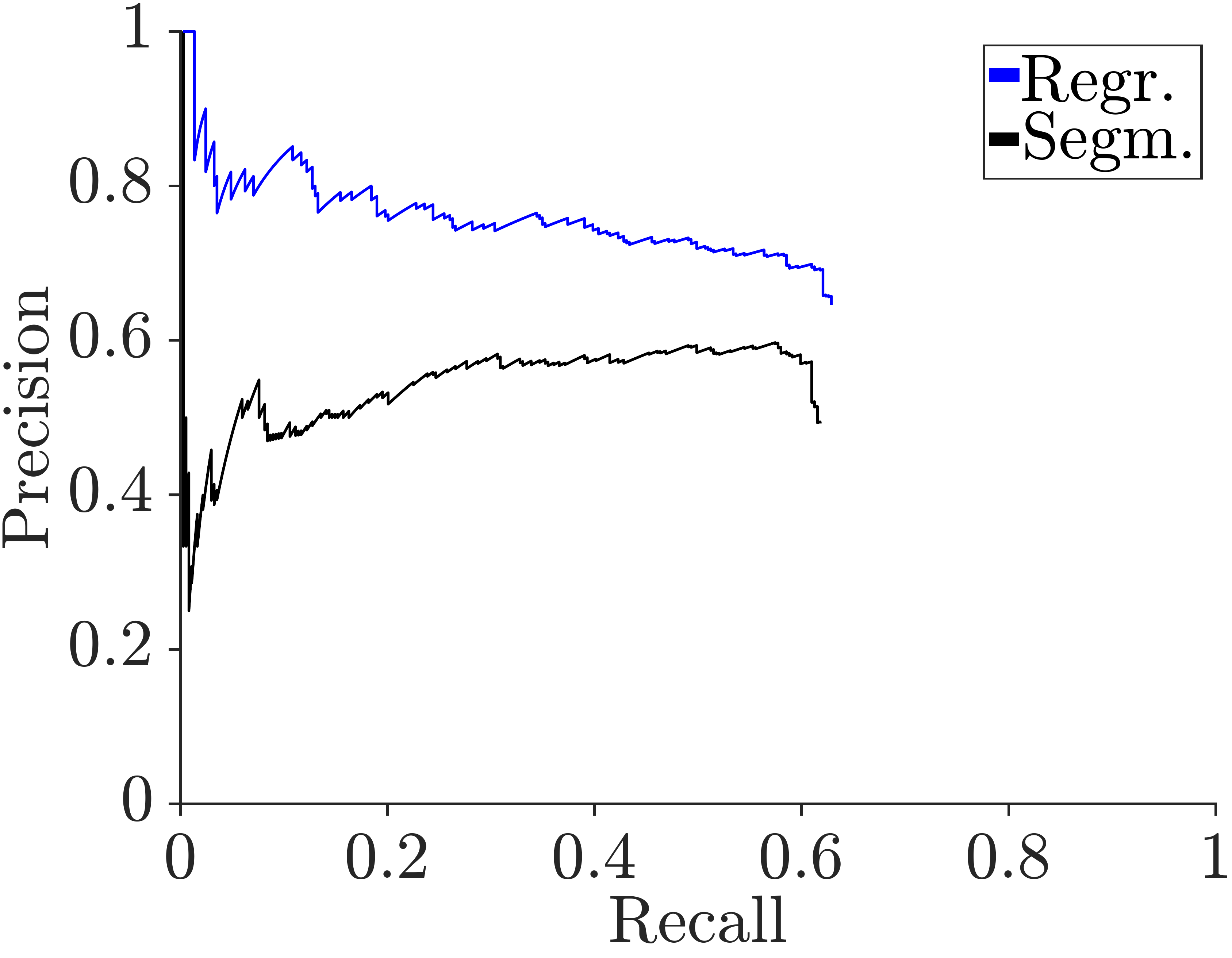}
\end{minipage}
  \caption{Precision-recall relationship of SEP localizations derived for three test image sets taken 4, 5 and \SI{8}{weeks} after sugar beet germination. The \dora{} data set additionally has a very high weed density. Acceptance ranges are \SI{6}{px} and \SI{18}{px}. The first baseline is abbreviated as SVM, the second as Segm. and our approach as Regr.}\label{fig:eval:pr_re}
\end{figure}

\paragraph*{Early Growth Stage}
The \anton{} sequence of \SI{100}{images} was recorded when most sugar beet plants were in a growth stage of two to four leaves and one plant's leaves rarely overlap.
The individuals are usually well separated by soil. 
Also, only few weeds are visible such that the ground truth contains \SI{166}{instances} of sugar beet and only \SI{10}{instances} of weed plants.
Fig. \ref{fig:eval:pr_re} shows large differences in precision and recall between the SVM baseline and the neural network based approaches.
However, the network approaches can not clearly be separated.
Tab. \ref{tab:map_ap} shows both neural network approaches performing very similar with about \mbox{\SI{80}{\%} AP}  and only \SI{5.9}{px} MAD.

\paragraph*{Weed-Affected}
The \dora{} sequence is an example for a more weed affected situation. 
The ground truth features \SI{330}{instances} of sugar beet and \SI{1427}{instances} of weed plants.
This high weed density can occur due to non-effective pest control or on the borders of a field.
Accordingly, the images contain a higher visual variance, as different plant species occur and leaves often overlap.
Here, the precision recall relationships in Fig. \ref{fig:eval:pr_re} separate the approaches better.
Our approach can handle the high number of weed plants better than the baselines for both distance thresholds.
As described in Tab. \ref{tab:map_ap}, the pose regression approach reaches as low as \SI{7.96}{px} in MAD, which is a relative improvement of \SI{5.9}{\%} over the segmentation network approach. 
More importantly, it performs a relative \SI{51,6}{\%} better in AP. 
The SVM baseline underperforms in this scenario.

\begin{table}[tp]
\centering
\begin{tabular}{cccc}
\hline
Dataset & Approach & AP & MAD $[\textrm{px}]$  \\ 
\hline
\anton{} 	& \svm{} 		& 0.515		& 10.12 \\ 
 			& \segm{} 		& \textbf{0.798} 	& \textbf{5.82}  \\ 
 			& \regr{} 		& 0.796 	& 5.90  \\ 
 \hline
\dora{} 	& \svm{} 	 	& 0.054 	& 10.58 \\ 
		 	& \segm{}	 	& 0.285 	& 8.43  \\ 
 			& \regr{}	 	& \textbf{0.432} 	&  \textbf{7.96} \\ 
\hline
\heinrich{} & \svm{} 		& -- 		& --  	\\ 
		 	& \segm{}	 	& 0.447 	& 10.98 \\ 
 			& \regr{} 	 	& \textbf{0.513}  	& \textbf{10.05} \\ 
\hline 
\end{tabular} 
\caption{Average Precision (AP) and Mean Accepted Distances (MAD) for the test image set comparison. \SI{1}{pixel} corresponds to about \SI{0.6}{mm} ground resolution. The first baseline is abbreviated as SVM, the second as Segm. and our approach as Regr.} \label{tab:map_ap}
\end{table}

\paragraph*{Late Growth Stage}
The \heinrich{} sequence was recorded at the end of the crop management phase. Pest control has been successful for our test field so that this sequence features \SI{369}{instances} of sugar beet and only \SI{5}{instances} of weed plants (see Fig.~\ref{fig:intro:samples}d for an example image).
The SVM baseline would need a separate training specifically for this growth stage. 
Also, the neural network baseline performed generally better in the other test sequences, for which reason we exclude the SVM for the late growth stage.
The results in Fig.~\ref{fig:eval:pr_re} show that the regression approach achieves higher precision rates.
Tab. \ref{tab:map_ap} shows the mean accepted distance errors where regression ranks again before segmentation.
It performs a relative \SI{14.7}{\%} better in AP and has a \SI{9.3}{\%} relative improvement in  mean accepted distance. 
The lower accuracy compared to the \anton{} sequence, with comparably low amounts of weed plants, could be explained by the older sugar beet plants containing less texture in their center.

\subsection{Landmark Reproducibility}
This experiment evaluates the pose regression SEP localization approach in the robotic use-case by comparing two mapping runs of the agricultural robot.
Effectively, this shows how system uncertainties influence the geo-referencing  and reproducibility of SEP landmarks.
We evaluate the long-term reliability by varying the time between two mapping runs, taking the data again from \cite{chebrolu2017agricultural}.

The pipeline to produce a metric SEP landmark map from the image detections is as follows:
A high-precision GNSS is employed for all mapping runs to assign a reference pose to every image.
Empirical evaluation of the GNSS showed that \SI{50}{\%} of position errors ranged below \SI{3.1}{mm} and \SI{95}{\%} below \SI{7.9}{mm}.
Additionally, we improve the estimated positions by averaging the GPS fixes during constant robot motion and fuse odometry measurements in an Unscented Kalman Filter.
We use the Kalman filter's pose estimations as the origin for a pinhole camera projection (neglecting ground unevenness).
As the image recording produces overlapping image regions, we merge landmarks in the metric map if they are closer than the GNSS system's \SI{95}{\%} error range.

The evaluation is then done  by matching the landmarks for minimal Euclidean distance in two overlaid metric landmark maps.
From the resulting error distance distribution we choose the $3\sigma$-range (about \SI{70}{mm} in both experiments) as an acceptance range for a match.
We show figures for precision and recall rates given this acceptance range and evaluate the continuous distance errors.

\begin{figure}[tb]
  \centering \includegraphics[width=0.48\textwidth]{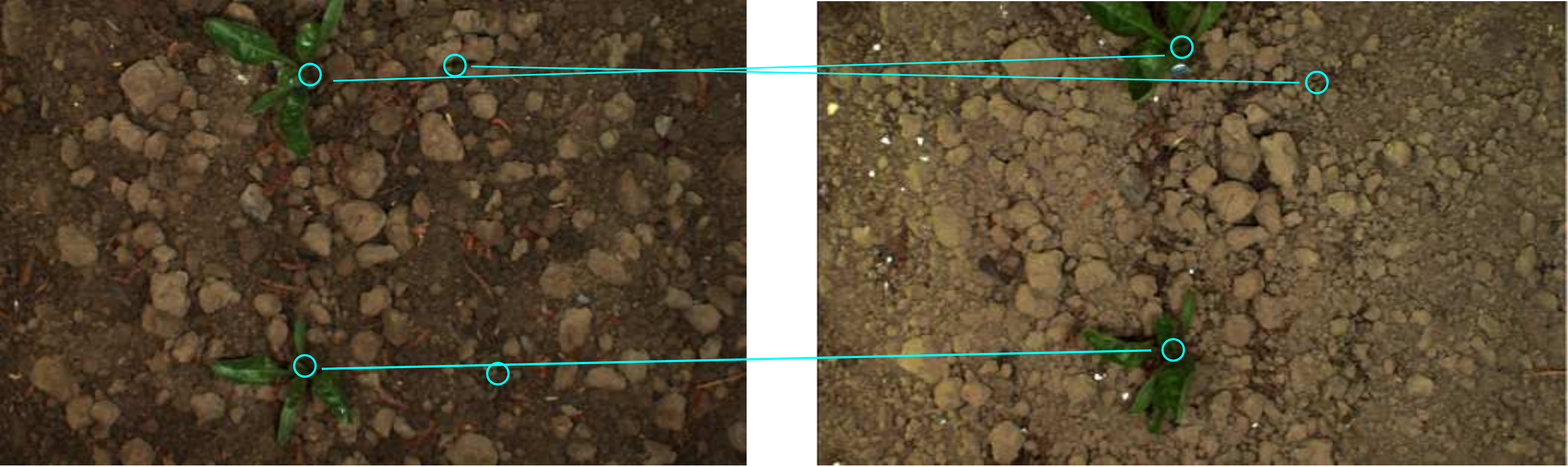}
  \caption{4 day comparison between geo-referenced images taken in the scope of the \ludwig{} (left) and \martha{} sequences. Detections from regression approach (circles). One weed has died and can not be matched. Only little growth is observable, the soil structure is still the same.}
  \label{fig:eval:map_compare_ludwig_martha_samples} 
\end{figure}

\addtolength{\textheight}{-.8cm}

\paragraph*{Evolution over 4 Days}
The \ludwig{} has been recorded \SI{4}{days} before the \martha{} sequence in the same traversing direction.
Still, Fig. \ref{fig:eval:map_compare_ludwig_martha_samples} gives an impression of how different perceptions of the field environment  can be even for a short time-span:
The ground might still show the same soil structure, but plants can hardly be associated by their appearance.
The leaf angle distribution can change and the earlier sequence generally shows more shimmering leaf surfaces due to more moisture on that day.
The error distributions in x and y directions have a very low bias ($\mu_x=\SI{1.2}{mm}, \sigma_x=\SI{16,7}{mm}, \mu_y=\SI{-2,9}{mm}, \sigma_y=\SI{13,3}{mm}$), which shows that the underlying pose-estimation is working properly.

\begin{figure}[tbp]
 \centering 
 \includegraphics[width=.47\textwidth]{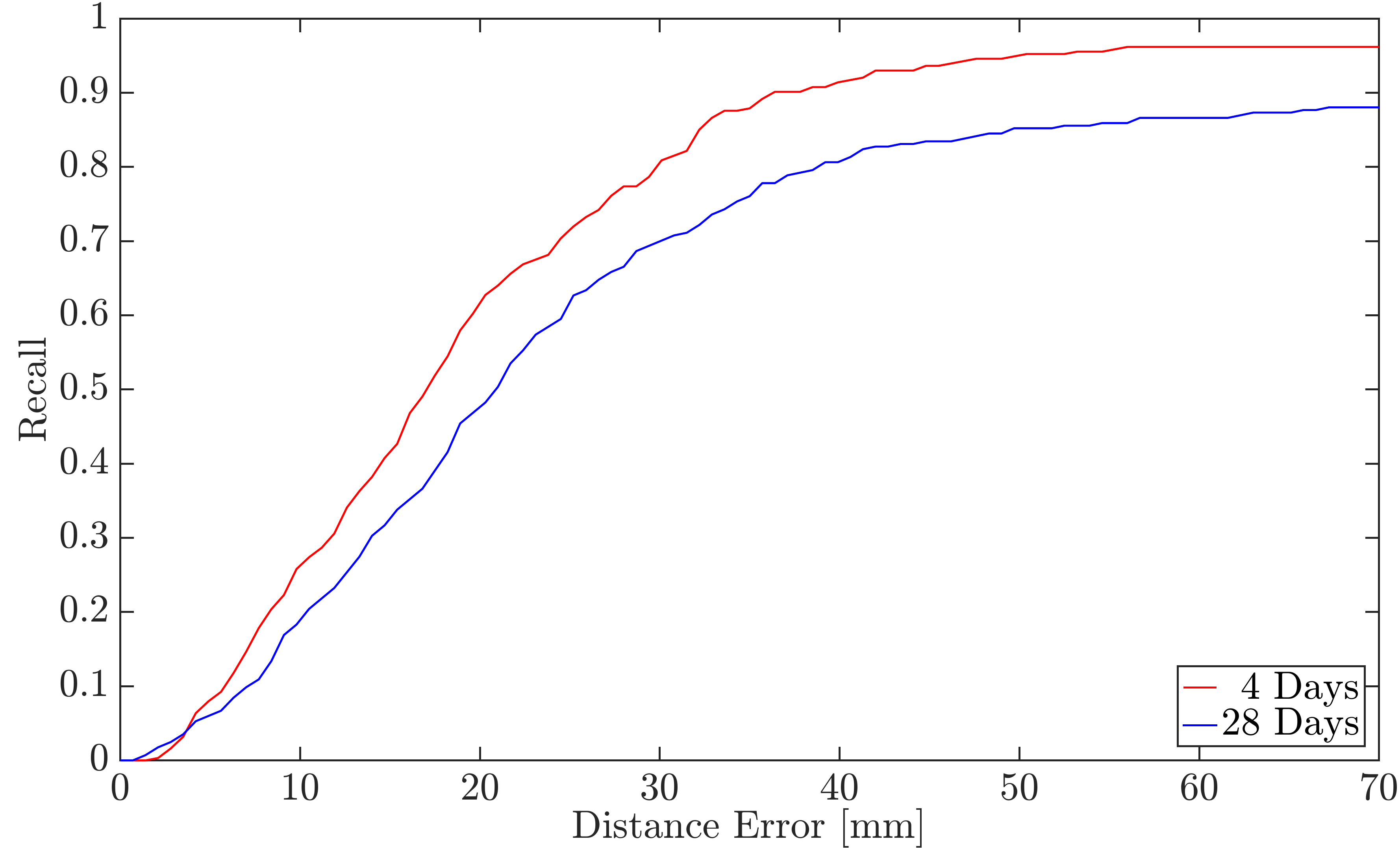}  
  \caption{Comparison of metric landmark maps for a \SI{4}{day} and \SI{28}{day} time span. The curves show the euclidean distance errors if every SEP landmark from an initial map is matched with the closest one in a later map.}\label{fig:eval:map_compare_integral}
\end{figure}

As a result, \SI{302}{plants} are matched successfully and only \SI{12}{outliers} are found.
Since the second mapping run finds \SI{345}{SEPs}, the corresponding recall rate is \SI{96.2}{\%} and precision equals \SI{87,5}{\%}.
The resulting errors with a mean of $\mu=\SI{18.41}{mm}$ are depicted in Fig.~\ref{fig:eval:map_compare_integral}.
Note that this is an upper bound, as it also contains the GNSS uncertainty.

\begin{figure}[tb]
  \centering \includegraphics[width=0.48\textwidth]{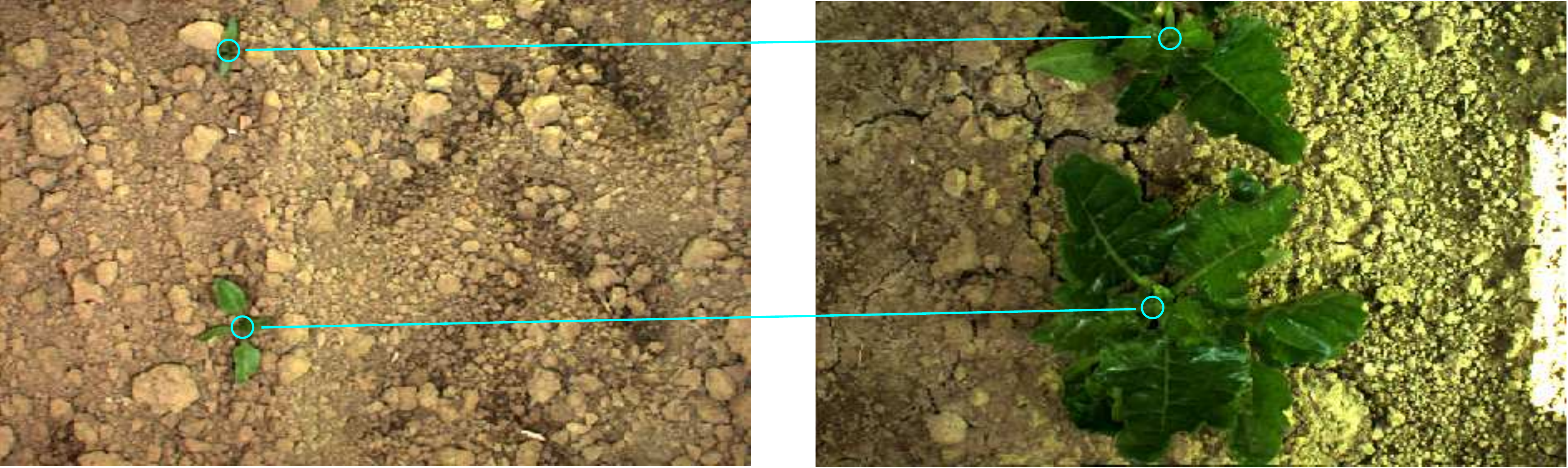}
  \caption{Geo-referenced images with a 28 day difference and severe changes in plant and soil appearance. Detections from regression approach (circles).}
  \label{fig:eval:map_compare_gustav_heinrich_samples}
\end{figure}

\paragraph*{Evolution over 28 Days} 

The \gustav{} sequence shows images from an early growth stage of two main leaves.
The \heinrich{} sequence has been recorded \SI{28}{days} later in opposite driving direction, shortly before the pre-harvesting phase of the field starts. Accordingly, the plants have grown a lot, overlapping plants are common and the leaves extend a lot more as can be seen in the comparison in Fig.  \ref{fig:eval:map_compare_gustav_heinrich_samples}.

The evaluation shows \SI{34}{outliers} for \SI{291}{SEPs} landmarks in the earlier map.
This corresponds to a recall rate of \SI{88,3}{\%} and precision rate of \SI{92,8}{\%}.
The distance errors of matched landmarks are shown in Fig.~\ref{fig:eval:map_compare_integral} ($\mu=\SI{20,8}{mm}$). 
This map comparison shows slightly higher mapping errors than the experiment  over \SI{4}{days} and at earlier growth stages did.
This corresponds to the image comparison experiment where later growth stages produced larger distance errors.
We conclude that the landmark matching over this long period shows good reliability and accuracy.

\section{CONCLUSIONS}\label{conclusions}
We present an approach for time-invariant plant localization on RGB and NIR image data.
Based on an FCNN and pose regression, it outperforms state-of-the-art approaches.
We show that the resulting localizations, when compared with an image ground truth, achieve high precision and recall rates, while mean distance errors can be expected to be as low as \SIrange{5}{11}{px}.
The image data set, ground truth annotations, and evaluation tools are publicly available.
For the agricultural robot scenario, we show that the SEP detections can be used as anonymous landmarks:
We were able to reproduce hundreds of landmarks with precision/recall rates of about \SI{90}{\%} and mean errors as low as \SI{20}{mm}.
For future work, we expect better regression network results if the likelihood decay parameter is chosen depending on the plant size.
This might increase accuracies for plants that offer less texture in the plant center, like sugar beets in late growth stages.
Our approach will also benefit from advances made in the field of FCNNs.
We will implement a robot localization approach based on our work, which enables an agricultural robot to navigate autonomously without the disadvantages of a high-precision GNSS.




\bibliography{ma_bib}
\bibliographystyle{plain}

\end{document}